# Cross-modal Representation Learning for Zero-shot Action Recognition

Chung-Ching Lin, Kevin Lin, Linjie Li, Lijuan Wang, Zicheng Liu
Microsoft
{chungching.lin, keli, lindsey.li, lijuanw, zliu}@microsoft.com

## Abstract

*We present a cross-modal Transformer-based framework, which jointly encodes video data and text labels for zero-shot action recognition (ZSAR). Our model employs a conceptually new pipeline by which visual representations are learned in conjunction with visual-semantic associations in an end-to-end manner. The model design provides a natural mechanism for visual and semantic representations to be learned in a shared knowledge space, whereby it encourages the learned visual embedding to be discriminative and more semantically consistent. In zero-shot inference, we devise a simple semantic transfer scheme that embeds semantic relatedness information between seen and unseen classes to composite unseen visual prototypes. Accordingly, the discriminative features in the visual structure could be preserved and exploited to alleviate the typical zero-shot issues of information loss, semantic gap, and the hubness problem. Under a rigorous zero-shot setting of not pre-training on additional datasets, the experiment results show our model considerably improves upon the state of the arts in ZSAR, reaching encouraging top-1 accuracy on UCF101, HMDB51, and ActivityNet benchmark datasets. Code will be made available.*[1]

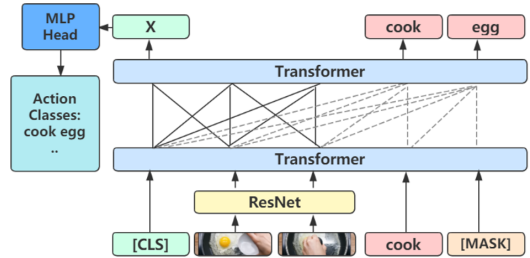

Figure 1. **ResT** is a cross-modal transformer network, which learns visual representations along with visual-semantic associations for ZSAR. In ResT, visual tokens attend to visual tokens (modality-specific attention); Word tokens attend to visual and text tokens on the left (cross-modal attention).

## 1. Introduction

Action recognition with supervised training is highly successful [9, 22, 37, 60, 61, 68], e.g., AssemblyNet++ [53] and X3D [17]. In comparison, zero-shot action recognition (ZSAR) generally lags behind because it requires a model to make meaningful inferences about unseen concepts, given only the data provided from seen training concepts and additional high-level semantic label information [57]. Addressing the general zero-shot challenges, e.g., distribution shift and semantic gap, recent ZSAR methods mainly exploit visual features extracted from off-the-shelf pretrained action recognition models and focus on studying a more robust visual-to-semantic mapping or learning a joint embedding space on which to project visual and semantic features. However, there are limitations in this typical framework.

First, visual features are usually acquired by pretrained action recognition models and remain unchanged during training (e.g., C3D [60] is employed in [4, 24, 43, 43, 65], I3D [9] is adopted in [21, 38, 46, 52]). Hence, they may not contain enough information for learning a fair representation [35]. Besides, when directly using pretrained action recognition models to extract visual features, the intrinsic zero-shot setting may not be preserved. The pretrained model acquires the knowledge of classes that should not be seen during training [52]. For example, I3D was trained on the ImageNet [13] and Kinetics [28] datasets. C3D was trained on the I380K and Sports-1M datasets [27]. Similar to Kinetics, between Sports-1M and UCF101 [55], they share 23 identical classes [16].

Second, action classes are usually manually labeled with limited descriptions of many complex actions, causing the semantic information is often imprecise or incomplete. For example, many action classes are only labeled by a noun (e.g., "uneven bars," "hula hoop"), or a single verb (e.g., "pour," "dive"). In contrast, videos, upon which the words are placed, are true visual reflections of various classes of actions. Visual observations are unstructured and complex. It follows, there is richer and more discriminative information in the visual feature space. Despite the visual and semantic spaces being connected, the visual discrimination after mapping or projection of the two spaces often shrinks to a certain extent [47, 73]. This affects the knowledge transferability from seen classes to unseen classes. Besides, as the projection is performed in high-dimensional embedding spaces, there inevitably exists the hubness problem [36, 50, 72], whereby some class prototypes appear to be the nearest neighbors of many irrelevant test instances.

---

[1] https://github.com/microsoft/ResT

The question is, *how can the visual and semantic spaces be bridged, while at the same time maintaining the visual discrimination for an effective knowledge transfer?*

In this paper, we study an end-to-end trainable cross-modal framework, named **ResT** (**Res**Net-**T**ransformer). ResT is able to associate both visual and semantic spaces, while preserving the descriptive and discriminative information implied in the visual embedding space. Considering the essence of ZSAR, we frame the problem in two ways: (1) Instead of using pretrained feature extractors, we attach a vanilla non-pretrained ResNet module to the Transformer to extract visual features, which is to ensure that no prior knowledge of unseen classes is acquired during training. (2) Contrary to normal practice, we integrate the learning of visual representations and visual-semantic associations into the same unified architecture. This model design provides a natural mechanism for visual and semantic representations to be learned in a shared knowledge space, which can bridge the semantic gap and encourage the learned visual embedding to be discriminative and more semantically consistent.

Considering the disconnection among source and target domains, which makes great difficulty in inferring their relationships via a coarse one-on-one nearest neighbor matching, we develop a simple semantic transfer scheme for zero-shot inference. The idea is to leverage the visual-semantic associations in ResT and to composite a visual prototype for each unseen class by embedding a combination of relevant information. Specifically, in light of the observations that human activities (e.g., play basketball) are composed of a series of simple actions (e.g., run, pass, jump and shoot) or are related to a set of partial elements of other complex activities, we posit that actions are implicitly compositional [1, 23, 39]. The semantic transfer scheme is thus formulated as a subgraph selection, based on the semantic relatedness distances of seen and unseen labels, to composite the visual prototype of each unseen class in visual space for the ZSAR task. With the transfer scheme, because the visual representations are not projected onto other spaces, the hubness problem is alleviated, and the visual distinction is preserved. Accordingly, the framework achieves good expandability in various unseen domains.

The present work follows a stricter, but more realistic zero-shot setting proposed by [5], where the set of training classes that overlap with test data are removed by a similarity threshold. The model is trained from scratch with random initial weights on one dataset and tested on three disjointed target datasets. No pre-training on auxiliary datasets is performed to ensure no prior knowledge of unseen classes acquired during training. Our approach employs the *inductive* configuration [4, 5, 24, 38, 72], where the test data is entirely unknown during training.

The main contributions of this paper are:

- Our approach, based on the described cross-modal framework, bridges the visual and semantic spaces while still maintaining the visual discrimination for an effective knowledge transfer. With a single trained model on the Kinetics dataset, our framework establishes new state-of-the-art ZSAR results on UCF101 [55], HMDB51 [33], and ActivityNet [7] benchmarks.

- We develop a simple yet effective semantic transfer scheme to composite unseen visual prototypes, with which ZSAR could be realized in the visual space to alleviate information loss and the hubness problem.

- Our approach has three nice properties: first, it is end-to-end trainable; second, it achieves a good accuracy–complexity trade-off; third, it offers the flexibility of utilizing different feature encoder backbones and is capable of cooperating with concurrent pretrained models for generalization.

## 2. Related Work

**Visual and Semantic Association.** In ZSAR, the visual-semantic mapping can be generally summarized in three main approaches.

First, a wide range of methods [19, 21, 42, 43, 48, 64–67, 71, 76] employ an indirect semantic-visual mapping. Both visual and semantic embeddings are projected onto a common intermediate space, and ZSAR is performed in the space. Wang and Chen [65] propose a two-stage framework to learn a latent embedding space and embed the semantic representations of unseen-class labels onto it via the guidance of the landmarks. Recently, Chen and Huang [10] recommend a method with human involvements to construct Elaborative Descriptions sentences from action class names using Wikipedia and Dictionary, and generate Elaborative Concepts of the objects in videos using WordNet. The method uses the BiT model [31] pretrained on ImageNet21k [13] to predict object probabilities, and leverages objects involved in seen and unseen classes for ZSAR.

Second, instead of indirect mapping, many methods [2, 4, 5, 24, 30, 34, 69, 70] have been proposed to project visual features into semantic embedding space and perform classifications directly in that space. Brattoli et al., [5] propose an end-to-end method that jointly optimizes the visual embeddings acquired by C3D or R(2+1)D [61] architectures and the semantic embeddings computed by the Word2Vec function. Third, the reverse mapping from semantic to visual embeddings is another option [38, 72] to alleviate the hubness problem. Zhang and Peng [72] propose to model the joint distribution over high-level video features and semantic knowledge via the Generative Adversarial Network (GAN), in which seen-to-unseen correlation is embedded to synthesize video features of unseen categories. Mandal et al. [38] adapt the conditional Wasserstein GAN with additional loss terms to synthesize unseen features for training an out-of-distribution detector. Unlike prior work using class labels to synthesize visual features via GANs, our model leverages the learned visual representations of seen classes to construct the unseen visual prototypes.

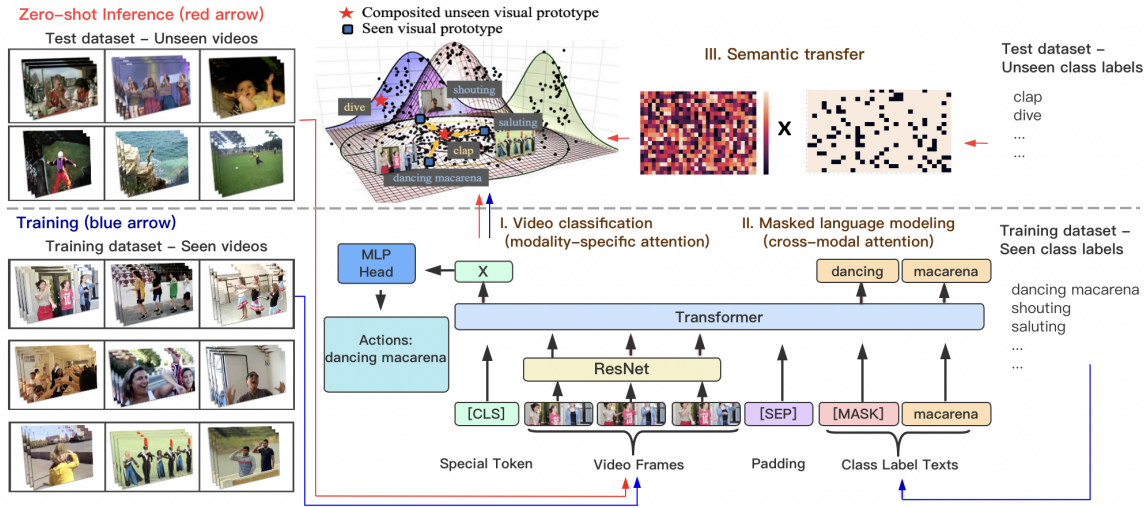

Figure 2. **Summary of the framework.** At the training stage, ResT jointly encodes video-text pairs in a single cross-modal Transformer to learn discriminative and more semantically consistent visual representations. At inference, the semantic transfer scheme embeds the semantic relatedness information between seen and unseen class labels to composite unseen visual prototypes. The model then takes a new observation with a single-modality input (unseen video) to produce its visual representation for zero-shot action classification. (Best viewed in color).

**Transformer Architecture.** Transformer architectures have shown exemplary performance on a wide range of applications, e.g., image recognition [15, 59], object detection [8, 11, 75], and visual-language tasks [18, 26, 51, 56]. Among the models, the most popular ones include ViT [15], VideoBERT [56], VIVO [26], and CLIP [49]. Based on a contrastive approach to learning image representations from texts, CLIP achieves amazing zero-shot transfer capabilities on many downstream tasks. CLIP is trained on a large corpus of 400 million image-text pairs. It takes 18 days to train a CLIP model with ResNet backbone on 592 units of V100 GPUs. In contrast, our model is trained from scratch on the Kinetics dataset, which consists of around 500 thousand clips. This underlines the differences in the scale of the CLIP and the proposed method. We aim to showcase the extensibility of a model trained on a dataset with limited data points to multiple disjointed test datasets.

## 3. Model

The conceptual diagram of our framework is shown in Figure 2. ResT consists of a primary task and an auxiliary task. To learn discriminative visual representations, the primary task is trained under action classification without semantic cues. This task is modality-specific, without cross-talk through attention between two modalities (described in Section 3.2.2). The goal is to learn richer and structure-preserving visual embeddings for effective knowledge transfer. The auxiliary task performs masked language modeling with visual cues. For a given pair of video data and text labels in the training dataset, this task aims to take both modalities into account to predict a masked class label. This task drives the network to align the visual and semantic content (described in Section 3.2.3). Together, the joint objective enables the model to produce more semantically consistent visual representations.

After the training, with a simple semantic transfer scheme, we use the prototypes of the seen classes to composite visual prototypes of the unseen classes in the learned visual space by reflecting their semantic relatedness in the semantic label embedding space. During testing, as the visual representation learning task is modality-specific, the model could take a new observation with a single modality input (video data) to produce its visual representation. The representation is then classified with the nearest neighbor search by evaluating its distance to the unseen visual prototypes that are composited by the seen visual prototypes.

### 3.1. Problem Definition

Let $S = \{v_i^s, y_i^s | v \in V, y \in L^s\}$ be the training set for seen classes, where $v_i^s$ is a video instance in $V$ and $y_i^s$ is the class label in $L^s = \{l_1^s, ..., l_\kappa^s\}$ with $\kappa$ seen classes. Given the set $S$, we train a model $P$ to learn visual representations $\{x_i, ..., x_\nu\} \in X$, where $x_i \in \mathbb{R}^D$ denotes the $D$-dimension embedding in $X$. Let $L^u = \{l_1^u, ..., l_\gamma^u\}$ be a set of $\gamma$ unseen class labels. $L^u \cap L^s = \varnothing$. In addition, $\Xi^s = \{\xi_1^s, ..., \xi_\kappa^s\}$ and $\Xi^u = \{\xi_1^u, ..., \xi_\gamma^u\}$ denote two sets of semantic embeddings corresponding to $L^s$ and $L^u$. Given a testing video instance $v_j^u$, the ZSAR problem is to estimate its label $y_j^u \in L^u$.

## 3.2. ResT Model Formulation

Our model consists of a frame-level feature encoder $\mathcal{F}$ and a cross-modal transformer $\mathcal{T}$. We use a vanilla 2D ResNet [25] as a base network to extract visual features because 2D models are typically more efficient than 3D models in terms of memory and runtime. The outputs of frame features are fed into the transformer network. To keep ResT computationally efficient, we take the flattened global features after the last average pooling layer in $\mathcal{F}$ as the visual features, which has been shown to work reasonably well on video understanding tasks [63]. To ensure the restriction of zero-shot setting is preserved, both feature encoder $\mathcal{F}$ and transformer $\mathcal{T}$ are initialized with random weights. During the training, both $\mathcal{F}$ and $\mathcal{T}$ are optimized together.

### 3.2.1 Model inputs

ResT takes a video-text pair $(I, S)$ as an input, where $I$ is a sequence of $T$ sampled frames $\{I_1, I_2, ..., I_T\}$ in a video and $S$ is the corresponding action label text. For a frame $I_t$ sampled at time $t$ in a video $v$, frame-level features $r_t$ are extracted by the feature extractor, where $r_t \in \mathbb{R}^p$ is a $p$-dimension vector. The cross-modal transformer $\mathcal{T}$ takes $(r, w)$ pairs as inputs, where $r = [r_1, r_2, ..., r_T]$ is the set of frame feature vectors and $w = [w_1, ..., w_{N_c}]$ is the sequence of word embeddings of the corresponding class label $S$ with $N_c$ number of words.

The transformer uses a constant dimension, D, through all layers. To ensure that both of the visual and word embeddings have the same dimension, a trainable linear projection layer is added to map each visual feature $r_t$ to the model dimension D. The sequence of the input features $z_0$ to the transformer has the form: $z_0 = [[\texttt{CLS}], r, w]$, where a special token $[\texttt{CLS}] = z_0^0$ is prepended to the sequence of input for visual representation learning in 3.2.2.

### 3.2.2 Visual representation learning

The primary task in ResT is to perform video classification without semantic clues. We incorporate a feature aggregator and a classifier into the final layer of the transformer to predict which action class the visual representation belongs to. Given a training dataset, this task is to train a model $P$ to learn the visual representation $x = P(z_0)$ with only visual cue $r$. Our transformer consists of $L$ layers. We denote $z_l$ as the output of $l^{\text{th}}$ layers and $ATT$ as the attention mask.

To exploit all possible temporal relationships, all visual feature tokens are used with bidirectional attention to jointly encode all visual features together ($ATT(r_h, r_k) = 1, \forall r_h, r_k$). As every visual feature token is able to attend to all other visual feature tokens, the model could leverage both long-term and short-term temporal relationships to learn a better representation. In addition, we set all visual features not to attend to any of the word tokens ($ATT(r_h, w_k) = 0, \forall r_h, w_k$). This is to encourage the model to dedicate itself to learning the visual representations and prevent the model from behaving unexpectedly when label text input is not available during testing.

Because a transformer encoder creates an embedding for each of its feature inputs, we add a 1-hidden-layer MLP, $f(\cdot)$, as a classification head to the output of the special token $[\texttt{CLS}]$, $z_L^0$, to obtain a unique visual representation. The classification head, $f(\cdot)$, is used to predict the final video classes with Softmax cross-entropy loss: $\mathcal{L}_{cls} = -E_{(r)\sim S} \log p(y^s|f(x))$. Only $z_L^0$ is used for classification during training, which forces the embedding of the $[\texttt{CLS}]$ token to aggregate and contextualize all frame-level feature embeddings. $z_L^0$ serves as a video representation $x$. $x = z_L^0$.

### 3.2.3 Visual-semantic association

The auxiliary task in ResT performs masked language modeling with visual cues. Given a pair of input $(r, w)$, this task attends over both modalities to predict the masked tokens, reconstructing the corresponding class label. As illustrated in Figure 1, unlike visual tokens that attend to visual tokens only, word tokens attend to all visual and word tokens on their left side. The task is designed to align visual-semantic concepts and serve as a bridge to connect the visual and semantic spaces.

We apply the Masked Token Loss (MTL) to the discrete token sequence $w$ for training. This is similar to the standard task of Masked Language Modeling (MLM) in BERT [14]. Each input token in $w$ is masked at random with a probability of 15%, but the minimum number of masked tokens is set to one to ensure MLM takes part in each iteration. The masked one, $w_\alpha$, is replaced with a special token $[\texttt{MASK}]$. The objective of the training is to predict these masked tokens based on their surrounding word tokens $w_{\setminus \alpha}$ and all visual feature tokens $r$ by minimizing the negative log-likelihood: $\mathcal{L}_{mtl} = -E_{(r,w)\sim S} \log p(w_\alpha|w_{\setminus \alpha}, r)$. This task leads the model to grasp the dependencies between visual and semantic contents, thus driving the network to align the visual and semantic concepts.

### 3.2.4 Training

The loss of $\mathcal{L}_{cls}$ is used to guide the model in learning discriminative visual representations, and the loss of $\mathcal{L}_{mtl}$ is used to associate visual and semantic representations. By optimizing both loss functions in the same network, the model is trained to learn a more semantically consistent visual representation. The final training objective is the sum of both losses: $\mathcal{L} = \mathcal{L}_{cls} + \omega_{mtl} \cdot \mathcal{L}_{mtl}$, where $\omega_{mtl}$ is the weighting of $\mathcal{L}_{mtl}$.

## 3.3. Zero-shot Action Recognition

After the training, we generate a visual prototype $\varphi_i^s$ for each seen class $l_i^s$: $\varphi_i^s = \frac{1}{N_i^s} \sum_{q=1}^{N_i^s} x_q^{s,i}$, $i = 1, ..., \kappa$, where $x_q^{s,i}$ is $q^{\text{th}}$ visual representation of seen class $l_i^s$, and $N_i^s$ is the total number of videos in the seen class $l_i^s$.

### 3.3.1 Semantic relatedness transfer

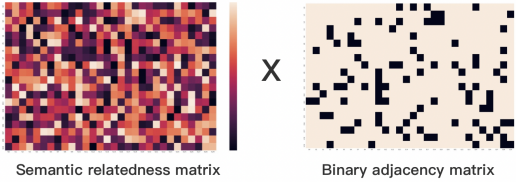

Figure 3. The examples of **semantic relatedness matrix** and **binary adjacency matrix** are created using subsets of label embeddings on ActivityNet and Kinetics. In the semantic relatedness matrix, brighter colors depict larger values. In the binary matrix, black color depicts the true value.

As ResT is designed to learn more semantically consistent visual representations, our supposition is that, the underlying embeddings share a co-occurrence relationship to some extent when the video-text pairs describe the same thing (e.g., the visual of "jump" & the text semantic of "jump") [74]. The idea of our transfer scheme is to use the outputs of the learned video-text pairs as anchors to build each unseen visual prototype, $\varphi_j^u$. We first obtain the most representative and distinctive bidirectional relationships between the seen and unseen class labels based on their semantic relatedness, and then embed the combination of this relevant information into the learned visual space to composite $\varphi_j^u$. A weighted sum is commonly used as a combination operator [6]. The proposed scheme is only applied in the *testing* phase, *without* employing any video instance from unseen classes.

Given a set of $\gamma$ unseen class labels $L^u = \{l_1^u, ..., l_\gamma^u\}$ and $\kappa$ seen class labels $L^s = \{l_1^s, ..., l_\kappa^s\}$ with their corresponding semantic embeddings $\Xi^u = \{\xi_1^u, ..., \xi_\gamma^u\}$ and $\Xi^s = \{\xi_1^s, ..., \xi_\kappa^s\}$ where $L^u \cap L^s = \emptyset$, our task is to link the best sets of the seen labels $L^s$ to the unseen labels $L^u$ subject to a set of constraints, such that we could obtain representative and distinctive composite visual prototypes for the unseen classes. This task can be formulated as a subgraph selection problem. Each class label represents a node. In this problem, for each of the unseen class nodes, we aim to construct a subgraph with possibly different sets of edges to the seen class nodes.

We first construct a semantic relatedness matrix $M \in \mathbb{R}^{\gamma \times \kappa}$ where each element $m_{j,i}$ in $M$ is defined as the semantic relatedness between $\xi_j^u$ and $\xi_i^s$. Here, $m_{j,i}$ is measured by cosine similarity: $m_{j,i} = cos(\xi_j, \xi_i)$. Let $A \in \mathbb{R}^{\gamma \times \kappa}$ denote a binary adjacency matrix, where each element $a_{j,i}$ in $A$ is a binary-valued variable that indicates whether the unseen class $l_j^u$ and the seen class $l_i^s$ are connected. We formulate this problem of subgraph selection as below. The goal is to find the assignments to the variables $a_{j,i}$ in $A$, which maximizes the objective function, with the space of the solutions bounded by a set of linear constraints.

$$\arg\max \sum M \odot A \quad (1)$$
$$\text{s.t. } a_{j,i} \leq \mathbb{1}_{l_i^s \in \mathcal{N}_{l_j^u}}, \quad (2)$$
$$a_{j,i} \leq \Lambda^r(j,i), \quad (3)$$
$$\sum_i a_{j,i} \leq \rho, \quad (4)$$

where $\odot$ represents the element-wise multiplication; $\mathbb{1}$ is an index function; $\mathcal{N}_{l_j^u}$ denotes $l_j^u$'s K-nearest neighbors according to a distance function $1 - m_{j,i}$; $\Lambda^r(j,i)$ is used to impose a relative distance constraint; $\rho$ is a threshold. We define $\Lambda^r(j,i)$ as:

$$\Lambda^r(j,i) = \begin{cases} 1, & \text{otherwise} \\ 0, & \min_{k \neq j}(\frac{1-m_{k,i}}{1-m_{j,i}}) \leq \vartheta, \end{cases} \quad (5)$$

where $\vartheta$ is a threshold. With the constraint in Eq. 2, only relatively close-related nodes are considered. Eq. 3 and Eq. 4 together promote the composited visual prototypes of unseen classes to be distinctive from one another.

We use an Integer Programming (IP) [12] to perform our search. With the IP solution $\hat{A}$, each row in $\hat{A}$ corresponds to an optimal subgraph. Specifically, in the $j^{\text{th}}$ row, all of the seen class $l_i^s$ with $a_{j,i} = 1$ are selected to composite the visual prototype of the unseen class $\hat{\varphi}_j^u$. To preserve the semantic relatedness, $\hat{\varphi}_j^u$ is calculated as follows:

$$\hat{\varphi}_j^u = \frac{1}{\sum a_{j,i} \cdot m_{j,i}} \sum_{i=1}^{\kappa} a_{j,i} \cdot m_{j,i} \cdot \varphi_i^s. \quad (6)$$

We follow a common protocol to compute the semantic embeddings of class labels [5, 52]. Given a class label with $N_c$ words $c_1, ..., c_{N_c}$, we adopt Word2Vec [41] to calculate the semantic embeddings $\xi$ of the class $j$ by averaging the word vectors to obtain a single fixed-size embedding vector: $\xi_j = \frac{1}{N_c} \sum_{k=1}^{N_c} w2v(c_k) \in \mathbb{R}^{300}$.

### 3.3.2 Zero-shot evaluation

With $\{\hat{\varphi}_j^u\}$, we can now perform the ZSAR in the learned space. For any unseen video instance $(v^u, r^u)$, we first input $r^u$ into the learned model $P$ to extract its visual representation $x^u$. $x^u = P(z_0^u)$, where $z_0^u = [[\texttt{CLS}], r^u, [\texttt{MASK}]]$. The zero-shot recognition of $v^u$ is achieved by evaluating its distance to the composite visual representations of the unseen classes. We assign the unseen label $y^* \in L^u$ to the unseen video instance $v^u$ as follows: $y^* = \arg\min_j d(x^u, \hat{\varphi}_j^u)$, where $d(\cdot)$ denotes cosine distance.

## 4. Experiments

### 4.1. Experimental Setup

**Datasets and training/evaluation protocol.** We train our model on the Kinetics dataset [28] and perform evaluations on three action recognition datasets: UCF101 [55], HMDB51 [33], and ActivityNet [7]. For a fair comparison, we adopt the same protocol introduced in [5] to remove a source category if its label is semantically identical to one of the target categories by a similarity threshold. This results in a subset of Kinetics 700 with 664 classes when classes are removed with respect to UCF ∪ HMDB, and a subset of 605 classes considering UCF ∪ HMDB ∪ ActivityNet [5]. We refer the reader to the E2E [5] paper for the detailed procedure. Our models are only trained once on the Kinetics dataset, without additional training on 50% of the target datasets. In the 0/50 (seen/unseen) split, we randomly select half classes from the test dataset, randomly generate 10 splits and report the averaged results.

**Implementation details.** We sample $T$ frames with stride 6 as a clip from each video at a random start time. Each frame is randomly cropped to a $224 \times 224$ patch. We design three models, named ResT_18, ResT_34, and ResT_101, where we replace the backbone of feature encoder $\mathcal{F}$ with ResNet-18 (512D), ResNet-34 (512D), and ResNet-101 (2048D). To maintain a similar magnitude of computation complexity, the corresponding sampled frames for three models are 16, 8, and 4 frames. We use one clip for training and 25 clips for testing per video. The transformer in ResT consists of 12 layers with a hidden size of 768D.

All our models are trained from scratch with random initial weights. We observe that directly optimizing the tasks in the encoder $\mathcal{F}$ and the transformer $\mathcal{T}$ with randomly initialized weights is not effective. Thus, we divide the training process into two stages: warm-up and joint-training. For the warm-up, we train only $\mathcal{F}$ for 150 epochs using SGD with a learning rate of 0.01 and weight decay of 0.0001. Then the whole pipeline is jointly trained for 50 epochs using AdamW with a learning rate of 0.00015 and weight decay of 0.05. The cosine learning rate scheduler is used on both stages. All experiments are performed on the Nvidia Tesla V100 GPUs. The mini-batch size is 16 clips per GPU. To train ResT focusing on visual representation learning, we set the MTL loss weight, $\omega_{mtl}$, to 0.5. For semantic relatedness transfer, the coefficients ($\vartheta$, $K$, $\rho$) are obtained with 5-fold cross-validation on the training class labels.

### 4.2. Main Results

We compare our approach with recent state-of-the-art *inductive* ZSAR methods. The experimental results on 50% class split, summarized in Table 1, show that our approach is competitive among these leading methods, achieving 54.7% and 39.3% top-1 accuracy on UCF101 and HMDB51 with ResT_18. We attribute the considerable gain in accuracy to the reduced information loss in our unified cross-modal framework, which is designed to exploit visual discriminations for effective knowledge transfer. The last two sets of results demonstrate that the performance could be further improved with deeper networks while keeping similar computational complexity in GFLOPs (shown in the ablations).

Table 1. **Comparison with recent state-of-the-art** on the 50% classes of UCF101 and HMDB51. The results are top-1 (%) with mean and standard deviation evaluated on *inductive setting*.

| Method | Pre VE[1] | SE[1] | UCF101 | HMDB51 |
|---|---|---|---|---|
| GA [43], WACV18 | C3D[2] [60] | W[1] | 17.3 ± 1.1 | 19.3 ± 2.1 |
| | | A[1] | 22.7 ± 1.2 | - |
| TARN [4], BMVC19 | C3D[2] [60] | W | 19.0 ± 2.3 | 19.5 ± 4.2 |
| | | A | 23.2 ± 2.9 | - |
| CEWGAN [38], CVPR19 | I3D[2] [9] | W | 26.9 ± 2.8 | 30.2 ± 2.7 |
| | | A | 38.3 ± 3.0 | - |
| TS-GCN [19], AAAI19 | Obj[1,3] | W | 34.2 ± 3.1 | 23.2 ± 3.0 |
| PS-GNN [20], PAMI20 | Obj[1,3] | W | 36.1 ± 4.8 | 25.9 ± 4.1 |
| E2E_C3D [5], CVPR20 | **None** | W | 43.8 | 24.7 |
| E2E_R(2+1)D [5] | | | 48.0 | 32.7 |
| DASZL [29], AAAI21 | TSM[2] [37] | A | 48.9 ± 5.8 | - |
| ER [10], ICCV21 | S[1]+ Obj[1,4] | ED[1] | 51.8 ± 2.9 | 35.3 ± 4.6 |
| ResT_18 | | | 54.7 ± 2.3 | 39.3 ± 3.5 |
| ResT_34 | **None** | W | 55.2 ± 3.0 | 40.6 ± 3.5 |
| **ResT_101** | | | **58.7 ± 3.3** | **41.1 ± 3.7** |

[1] Visual embedding obtained by pretrained models (Pre VE); Semantic embedding (SE); Objects (Obj); Spatial features (S); Word embedding (W); Attributes (A); Elaborated descriptions by Wiki/Diction./WordNet (ED).
[2] Visual features from pretrained action recognition models: C3D [60] (trained on I380K [60] and Sports-1M [27]); I3D [9] (ImageNet [13] and Kinetics); TSM [37] (ImageNet [13] and Kinetics [28]).
[3] Object scores by GoogLeNet [58] (12,988-class ImageNet Shuffle [40]).
[4] Spatial features and object scores obtained by Big transfer model (BiT) [31] (ImageNet/ImageNet21K [13]).

Table 2. **Comparison with E2E [5] on all test classes (0/100)** in UCF101, HMDB51, and ActivityNet datasets.

| Method | UCF101 | | HMDB51 | | ActivityNet | |
|---|---|---|---|---|---|---|
| | 0/50 | 0/100 | 0/50 | 0/100 | 0/50 | 0/100 |
| E2E_R(2+1)D [5] | 44.1 | 35.3 | 29.8 | 24.8 | 26.6 | 20.0 |
| ResT_18 | 50.9 | 41.2 | 37.6 | 30.6 | 29.2 | 23.0 |
| ResT_101 | **55.9** | **46.7** | **40.8** | **34.4** | **32.5** | **26.3** |

[1] Both methods train on a subset of Kinetics with **605 classes**. The subset is the result of removing all classes whose distance to any class in UCF101 ∪ HMDB51 ∪ ActivityNet is smaller than a threshold.

Due to the domain shift issue, few studies have explored the more realistic cross dataset configuration, where the model is trained on one independent dataset and is evaluated on other disjoint datasets. As both E2E [5] and the proposed method adopt this configuration, we are able to perform further comparisons on 0/100 (seen/unseen) full dataset test. In Table 2, we compare our models with the best-performing model of E2E, E2E_R(2+1)D. Overall, our baseline ResT_18 achieves noticeable improvements on three datasets (e.g., 5.9% absolute gains in top-1 accuracy on 0/100 UCF101).

Next, we visualize a random subset of testing video in-

Table 3. **Transformer model**

(a) **Importance of the Transformer architecture**

| SE | ResT | Transfer | 0/50 | 0/100 |
|---|---|---|---|---|
| RNN | | ✓ | 28.2 | 20.5 |
| LSTM | | ✓ | 29.7 | 21.8 |
| Transformer | ✓ | ✓ | 54.7 | 44.3 |

(b) **Transformer model design**

| CLS | MLM | Transfer | 0/50 | 0/100 |
|---|---|---|---|---|
| ✓ | | | 47.4 | 40.0 |
| | ✓ | | 45.0 | 38.7 |
| ✓ | ✓ | | 48.9 | 41.5 |
| ✓ | ✓ | ✓ | 54.7 | 44.3 |

(c) **Attention scheme**

| | CLS | MLM | 0/50 | 0/100 |
|---|---|---|---|---|
| Attention | Cross | Cross | 39.0 | 30.7 |
| | Single | Cross | 54.7 | 44.3 |

Table 4. **Feature extractor backbone**

(a) **Performance of different feature encoder backbones**

| Network | Frames | GFLOPs × clips | 0/50 | 0/100 |
|---|---|---|---|---|
| VGG16 | 16 | 248× 25 | 52.5 | 42.4 |
| MobileNetV2 | 16 | 6.9 × 25 | 50.6 | 38.5 |
| GoogleNet | 16 | 25.9 × 25 | 51.6 | 40.3 |
| R(2+1)D Archi. | 16 | 41.2 × 25 | 59.5 | 48.6 |

(b) **Accuracy-complexity trade-off**

| Network | Frames | GFLOPs × clips | 0/50 | 0/100 |
|---|---|---|---|---|
| E2E_R(2+1)D | 16 | 40.8 × 25 | 48.0 | 37.6 |
| E2E_R(2+1)D | 16 | 40.8 × 1 | 43.0 | 35.1 |
| ER_S+Obj | 8 | 70.2 × 1 | 51.8 | - |
| ResT_18 | 16 | 30.8 × 25 | 54.7 | 44.3 |
| ResT_18 | 16 | 30.8 × 1 | 54.0 | 43.5 |
| ResT_34 | 8 | 30.3 × 1 | 54.7 | 43.8 |
| ResT_101 | 4 | 32.0 × 1 | 57.0 | 47.7 |

Table 5. **ZSAR in semantic or visual space**

| SE | ResT | Transfer | Space | 0/50 | 0/100 |
|---|---|---|---|---|---|
| W2V | ✓ | | S | 48.9 | 41.5 |
| GloVe | ✓ | | S | 48.6 | 40.8 |
| S2V | ✓ | | S | 50.9 | 40.7 |
| W2V | ✓ | ✓ | V | 54.7 | 44.3 |
| Glove | ✓ | ✓ | V | 52.6 | 41.8 |
| S2V | ✓ | ✓ | V | 55.9 | 42.8 |

Table 6. **with and w/o pre-trained features and end-to-end training**

| VE | pre-trained | e2e | UCF | HMDB |
|---|---|---|---|---|
| C3D | ✓ | | 40.5 | 26.5 |
| C3D | ✓ | ✓ | 52.7 | 37.1 |
| Object | ✓ | | 57.3 | 40.5 |
| ResT_101 | | ✓ | 58.7 | 41.1 |

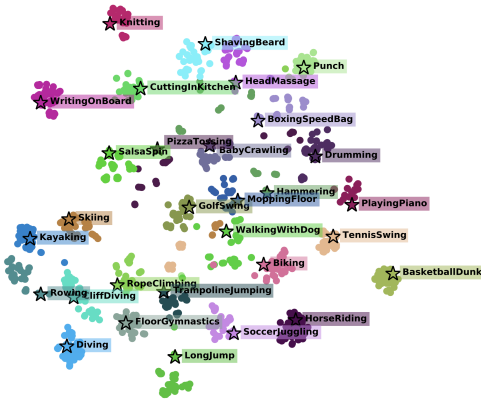

Figure 4. **Visualization** of data points from 30 random unseen classes on UCF101 with the learned visual representations $X$. Each color corresponds to one unseen class. The stars denote the unseen visual prototypes, which are composited from the seen visual prototypes by the proposed semantics transfer scheme.

stances on UCF101 in Figure 4, where each color corresponds to an unseen class label. It can be seen from the t-SNE plot that most of the test samples belonging to the same classes are in the vicinity, which shows that our model could learn discriminative visual representations. Overall, we observe that while the unseen composite prototypes deviate from the centroid of the corresponding instances, they are still distinctive enough to classify various unseen classes. Qualitatively, the results provide supports for the effectiveness of the proposed semantic transfer scheme. One example of the unseen prototypes composited from seen classes is that "clap" is composited by "shouting," "singing," "saluting," and "dancing macarena" with semantic relatedness $m_{j,i} = 0.42, 0.48, 0.50$ and $0.50$, respectively (Figure 2).

### 4.3. Ablation Study

For a controlled evaluation, unless noted otherwise, the ablation results are performed using the same training recipe with randomly initialized weights in ResT_18 (Kinetics 664) model on the UCF101 dataset.

**Importance of the Transformer architecture.** We first replace the Transformer with RNN and LSTM architectures to validate the rationale of choosing a Transformer as our main architecture. In the experiment with a single RNN/LSTM architecture taking both modalities, we observe the models converge to a trivial solution as the models rely on text modal for classifying actions. Thus, we experiment with a two-stream approach [73]. Table 3(a) shows that the RNN/LSTM models perform substantially worse than the Transformer model. The difference is that, when compared to RNN/LSTM, a great advantage in Transformer architecture for our task is the controllable attention mechanism $ATT(\cdot)$ and the special token [CLS]. Controlling the amount of information obtained through the attention and from the specific tokens, our model design encourages the visual and semantic representations to be learned in a shared knowledge space while still preserving visual distinctions.

**Impact of CLS and MLM components.** The ResT consists of two major components: classification (CLS) and masked language modeling (MLM). To evaluate the importance of each component, we train several models, turning them on and off. The proposed transfer scheme is not applicable to CLS-only or MLM-only models as two spaces are not associated. In the evaluation, a seen label $\tilde{y}^s$ is first assigned to unseen $v^u$ by a cosine distance, and ZSAR is then performed by the nearest neighbor (NN) search between seen $\tilde{y}^s$ and unseen $\{y^u\}$. Table 3(b) shows that CLS-only and MLM-only models obtain inferior results. The auxiliary MLM is served as a bridge to align semantic and visual concepts. It is designed not to directly involve in the video classification task. Although the direct benefit from MLM is modest, the inclusion of MLM is essential because it relates two spaces and enables the application of the transfer scheme. Consequently, the framework could exploit the visual capacity to a greater extent. The last row shows that our design choices of CLS, MLM, and transfer improve results and lead to higher Top-1 scores.

**Attention mechanism.** The way contextual information is explored via the attention mechanism is a key factor of the proposed method. To validate our design choice, we experiment with a version of ResT_18 with cross-modality atten-tion in both CLS and MLM, where both visual and word tokens attend to visual and text contexts. In Table 3(c), We notice it leads to substantial performance degradation. After 20 epochs, the training accuracy reaches 100%. This is because CLS relies solely on semantic cues for the classification task, making it fail to learn informative visual representations. The benefit of single-modality attention in CLS is that the task is constrained in the same modality, rather than absorbing the complementary information from the other modality. It also prevents the model from performing unexpectedly when it takes only a single-modality video input during inference.

**Performance of different feature encoder backbones.** We experiment with variants, replacing the ResNet feature encoder with different 2D/3D CNN backbones: VGG16, MobileNetV2, GoogleNet and R(2+1)D. No pretrained weights are used in all variants. Table 4(a) reports the computational complexity and accuracy evaluated on 25 clips. Among the 2D backbones, the model with MobileNetV2 achieves reasonable accuracy while using much less computation cost. We can observe that the model with a 3D CNN backbone (R(2+1)D) achieves higher accuracy at the cost of higher computational complexity.

**Accuracy-complexity trade-off.** Table 4(b) compares the accuracy and computation cost with respect to different numbers of clips in inference. The results of another end-to-end training method, E2E_R(2+1)D, and the SOTA, ER [10], are included for reference. At inference, we randomly sample $N_e$ clips (e.g., $N_e$=1 and 25) from each video and average these $N_e$ clip predictions to obtain the final results. The result shows that using one clip in ResT_18 is only within 1% loss in accuracy compared to using 25 clips. We hypothesize that, with discriminative visual information, one clip might contain sufficient information for recognizing actions. We also study the effect of using fewer frames for training. As stated in implementation details, to achieve computational efficiency, the number of frames sampled for training and testing in ResT_18, ResT_34, and ResT_101 are 16, 8, and 4 frames. In Table 4(b), we observe that training on a deeper network with fewer frames gives a better trade-off between computation cost and accuracy.

**ZSAR in different spaces.** In this experiment, we disable the transfer scheme and map each representation of unseen instance to the semantic space using different word embedding approaches (e.g., Word2Vec, GloVe [45], Sent2Vec [44]). We also evaluate the proposed semantic transfer scheme using different word embeddings to perform ZSAR in the visual space. As shown in Table 5, it can be seen that the classification accuracy in the semantic space is noticeably lower than that in the visual space. This could be due to the information loss in the mapping process. The semantic information is generally incomplete and less discriminative; thus, the model's generalization capability is constrained.

In contrast, our model achieves consistently higher accuracy with the proposed semantic transfer scheme using different word embeddings. We hypothesize that it is mostly because the discriminative property in the visual space is substantial. Also, because the design of the transfer scheme is to embed a combination of the most representative and distinctive information, the proposed framework is thus less prone to the hubness problem and the bias with NN search.

**Generalization.** While we propose a framework in which the model is not pretrained on additional datasets, the proposed model is flexible and capable of cooperating with other pretrained models. In this experiment, instead of integrating a ResNet module to encode visual features, we consider three cases to showcase the generalization of the proposed framework: *(i)* take pretrained visual features by a C3D model; *(ii)* integrate a pretrained C3D module and perform end-to-end training with our model; *(iii)* take pretrained object features [3]. In Table 6, the first two rows of results show our model is able to cooperate with concurrent models, and the end-to-end training allows our model to finetune task-specific features for performance enhancement. For example, compared to directly taking pretrained C3D features, incorporating the pretrained C3D model to train end-to-end with our model boosts the top-1 accuracy on UCF from 40.5% to 52.7%. In the object experiments, considering some unseen classes (e.g., "balance beam") are simply named for primary objects in videos, we only take object features into the model, but not object labels. This is to prevent the model achieving high accuracy by matching object names instead of recognizing actions. Table 6 shows that our model is able to handle contextual information in object features and make the classification of actions relatively effective. In sum, beyond the proposed e2e training scheme, it is also suitable to adopt our framework to explore other ZSAR solutions. More details and our limitations are discussed in the supplementary.

## 5. Conclusions

We have explored a cross-modal transformer-based framework that could establish effective knowledge transfer for the ZSAR task, where the distributional shift, semantic gap, and hubness problem exist and affect the way by which many heretofore existing methods perform. The competitiveness of our method on three benchmark datasets suggests that preserving the discriminative capacity in the visual embedding space can be a core factor for success in ZSAR. Comprehensive ablation studies indicate several key factors and advantages of the proposed model, including Transformer model design, computational efficiency, and flexibility of e2e training with various feature encoder backbones. Further, the proposed model is capable of cooperating with concurrent pretrained models for generalization.

# References

[1] Jake K Aggarwal and Michael S Ryoo. Human activity analysis: A review. *ACM Computing Surveys (CSUR)*, 43(3):1–43, 2011. 2

[2] Ioannis Alexiou, Tao Xiang, and Shaogang Gong. Exploring synonyms as context in zero-shot action recognition. In *2016 IEEE International Conference on Image Processing (ICIP)*, pages 4190–4194. IEEE, 2016. 2

[3] Peter Anderson, Xiaodong He, Chris Buehler, Damien Teney, Mark Johnson, Stephen Gould, and Lei Zhang. Bottom-up and top-down attention for image captioning and visual question answering. In *Proceedings of the IEEE conference on computer vision and pattern recognition*, pages 6077–6086, 2018. 8, 17

[4] Mina Bishay, Georgios Zoumpourlis, and Ioannis Patras. Tarn: Temporal attentive relation network for few-shot and zero-shot action recognition. In *BMVC*, 2019. 1, 2, 6

[5] Biagio Brattoli, Joseph Tighe, Fedor Zhdanov, Pietro Perona, and Krzysztof Chalupka. Rethinking zero-shot video classification: End-to-end training for realistic applications. In *Proceedings of the IEEE/CVF Conference on Computer Vision and Pattern Recognition*, pages 4613–4623, 2020. 2, 5, 6, 15

[6] Wieland Brendel and Matthias Bethge. Approximating cnns with bag-of-local-features models works surprisingly well on imagenet. In *International Conference on Learning Representations*, 2018. 5

[7] Fabian Caba Heilbron, Victor Escorcia, Bernard Ghanem, and Juan Carlos Niebles. Activitynet: A large-scale video benchmark for human activity understanding. In *Proceedings of the ieee conference on computer vision and pattern recognition*, pages 961–970, 2015. 2, 6, 17

[8] Nicolas Carion, Francisco Massa, Gabriel Synnaeve, Nicolas Usunier, Alexander Kirillov, and Sergey Zagoruyko. End-to-end object detection with transformers. In *European Conference on Computer Vision*, pages 213–229. Springer, 2020. 3

[9] Joao Carreira and Andrew Zisserman. Quo vadis, action recognition? a new model and the kinetics dataset. In *proceedings of the IEEE Conference on Computer Vision and Pattern Recognition*, pages 6299–6308, 2017. 1, 6

[10] Shizhe Chen and Dong Huang. Elaborative rehearsal for zero-shot action recognition. *International Journal of Computer Vision*, 2021. 2, 6, 8

[11] Ting Chen, Saurabh Saxena, Lala Li, David J Fleet, and Geoffrey Hinton. Pix2seq: A language modeling framework for object detection. *arXiv preprint arXiv:2109.10852*, 2021. 3

[12] Michele Conforti, Gérard Cornuéjols, Giacomo Zambelli, et al. *Integer programming*, volume 271. Springer, 2014. 5

[13] Jia Deng, Wei Dong, Richard Socher, Li-Jia Li, Kai Li, and Li Fei-Fei. Imagenet: A large-scale hierarchical image database. In *2009 IEEE conference on computer vision and pattern recognition*, pages 248–255. Ieee, 2009. 1, 2, 6

[14] Jacob Devlin, Ming-Wei Chang, Kenton Lee, and Kristina Toutanova. Bert: Pre-training of deep bidirectional transformers for language understanding. *arXiv preprint arXiv:1810.04805*, 2018. 4

[15] Alexey Dosovitskiy, Lucas Beyer, Alexander Kolesnikov, Dirk Weissenborn, Xiaohua Zhai, Thomas Unterthiner, Mostafa Dehghani, Matthias Minderer, Georg Heigold, Sylvain Gelly, et al. An image is worth 16x16 words: Transformers for image recognition at scale. *arXiv preprint arXiv:2010.11929*, 2020. 3

[16] Valter Estevam, Helio Pedrini, and David Menotti. Zero-shot action recognition in videos: A survey. *Neurocomputing*, 439:159–175, 2021. 1

[17] Christoph Feichtenhofer. X3d: Expanding architectures for efficient video recognition. *arXiv preprint arXiv:2004.04730*, 2020. 1

[18] Valentin Gabeur, Chen Sun, Karteek Alahari, and Cordelia Schmid. Multi-modal transformer for video retrieval. In *European Conference on Computer Vision (ECCV)*, volume 5. Springer, 2020. 3

[19] Junyu Gao, Tianzhu Zhang, and Changsheng Xu. I know the relationships: Zero-shot action recognition via two-stream graph convolutional networks and knowledge graphs. In *Proceedings of the AAAI conference on artificial intelligence*, volume 33, pages 8303–8311, 2019. 2, 6

[20] Junyu Gao, Tianzhu Zhang, and Changsheng Xu. Learning to model relationships for zero-shot video classification. *IEEE transactions on pattern analysis and machine intelligence*, 2020. 6

[21] Pallabi Ghosh, Nirat Saini, Larry S Davis, and Abhinav Shrivastava. All about knowledge graphs for actions. *arXiv preprint arXiv:2008.12432*, 2020. 1, 2

[22] Rohit Girdhar, Joao Carreira, Carl Doersch, and Andrew Zisserman. Video action transformer network. In *Proceedings of the IEEE/CVF Conference on Computer Vision and Pattern Recognition*, pages 244–253, 2019. 1

[23] Rohit Girdhar and Deva Ramanan. Cater: A diagnostic dataset for compositional actions & temporal reasoning. In *International Conference on Learning Representations*, 2019. 2

[24] Meera Hahn, Andrew Silva, and James M Rehg. Action2vec: A crossmodal embedding approach to action learning. *arXiv preprint arXiv:1901.00484*, 2019. 1, 2

[25] Kaiming He, Xiangyu Zhang, Shaoqing Ren, and Jian Sun. Deep residual learning for image recognition. In *Proceedings of the IEEE conference on computer vision and pattern recognition*, pages 770–778, 2016. 4

[26] Xiaowei Hu, Xi Yin, Kevin Lin, Lijuan Wang, Lei Zhang, Jianfeng Gao, and Zicheng Liu. Vivo: Surpassing human performance in novel object captioning with visual vocabulary pre-training. In *Proceedings of the AAAI conference on artificial intelligence*, 2021. 3

[27] Andrej Karpathy, George Toderici, Sanketh Shetty, Thomas Leung, Rahul Sukthankar, and Li Fei-Fei. Large-scale video classification with convolutional neural networks. In *Proceedings of the IEEE conference on Computer Vision and Pattern Recognition*, pages 1725–1732, 2014. 1, 6

[28] Will Kay, Joao Carreira, Karen Simonyan, Brian Zhang, Chloe Hillier, Sudheendra Vijayanarasimhan, Fabio Viola,

Tim Green, Trevor Back, Paul Natsev, et al. The kinetics human action video dataset. *arXiv preprint arXiv:1705.06950*, 2017. 1, 6, 12, 15

[29] Tae Soo Kim, Jonathan D Jones, Michael Peven, Zihao Xiao, Jin Bai, Yi Zhang, Weichao Qiu, Alan Yuille, and Gregory D Hager. Daszl: Dynamic action signatures for zero-shot learning. In *Proceedings of the AAAI conference on artificial intelligence*, 2021. 6

[30] Elyor Kodirov, Tao Xiang, Zhenyong Fu, and Shaogang Gong. Unsupervised domain adaptation for zero-shot learning. In *Proceedings of the IEEE international conference on computer vision*, pages 2452–2460, 2015. 2

[31] Alexander Kolesnikov, Lucas Beyer, Xiaohua Zhai, Joan Puigcerver, Jessica Yung, Sylvain Gelly, and Neil Houlsby. Big transfer (bit): General visual representation learning. In *Computer Vision–ECCV 2020: 16th European Conference, Glasgow, UK, August 23–28, 2020, Proceedings, Part V 16*, pages 491–507. Springer, 2020. 2, 6

[32] Ranjay Krishna, Yuke Zhu, Oliver Groth, Justin Johnson, Kenji Hata, Joshua Kravitz, Stephanie Chen, Yannis Kalantidis, Li-Jia Li, David A Shamma, Michael Bernstein, and Li Fei-Fei. Visual genome: Connecting language and vision using crowdsourced dense image annotations. 2016. 17

[33] Hildegard Kuehne, Hueihan Jhuang, Estíbaliz Garrote, Tomaso Poggio, and Thomas Serre. Hmdb: a large video database for human motion recognition. In *2011 International Conference on Computer Vision*, pages 2556–2563. IEEE, 2011. 2, 6, 17

[34] Yikang Li, Sheng-hung Hu, and Baoxin Li. Recognizing unseen actions in a domain-adapted embedding space. In *2016 IEEE International Conference on Image Processing (ICIP)*, pages 4195–4199. IEEE, 2016. 2

[35] Yanan Li and Donghui Wang. Joint learning of attended zero-shot features and visual-semantic mapping. In *BMVC*, 2019. 1

[36] Changsu Liao, Li Su, Wegang Zhang, and Qingming Huang. Semantic manifold alignment in visual feature space for zero-shot learning. In *2018 IEEE International Conference on Multimedia and Expo (ICME)*, pages 1–6. IEEE, 2018. 2

[37] Ji Lin, Chuang Gan, and Song Han. Tsm: Temporal shift module for efficient video understanding. In *Proceedings of the IEEE International Conference on Computer Vision*, pages 7083–7093, 2019. 1, 6

[38] Devraj Mandal, Sanath Narayan, Sai Kumar Dwivedi, Vikram Gupta, Shuaib Ahmed, Fahad Shahbaz Khan, and Ling Shao. Out-of-distribution detection for generalized zero-shot action recognition. In *Proceedings of the IEEE/CVF Conference on Computer Vision and Pattern Recognition*, pages 9985–9993, 2019. 1, 2, 3, 6

[39] Joanna Materzynska, Tete Xiao, Roei Herzig, Huijuan Xu, Xiaolong Wang, and Trevor Darrell. Something-else: Compositional action recognition with spatial-temporal interaction networks. In *Proceedings of the IEEE/CVF Conference on Computer Vision and Pattern Recognition*, pages 1049–1059, 2020. 2

[40] Pascal Mettes, Dennis C Koelma, and Cees GM Snoek. The imagenet shuffle: Reorganized pre-training for video event detection. In *Proceedings of the 2016 ACM on International Conference on Multimedia Retrieval*, pages 175–182, 2016. 6

[41] Tomas Mikolov, Kai Chen, Greg Corrado, and Jeffrey Dean. Efficient estimation of word representations in vector space. *arXiv preprint arXiv:1301.3781*, 2013. 5

[42] Ashish Mishra, Anubha Pandey, and Hema A Murthy. Zero-shot learning for action recognition using synthesized features. *Neurocomputing*, 390:117–130, 2020. 2

[43] Ashish Mishra, Vinay Kumar Verma, M Shiva Krishna Reddy, S Arulkumar, Piyush Rai, and Anurag Mittal. A generative approach to zero-shot and few-shot action recognition. In *2018 IEEE Winter Conference on Applications of Computer Vision (WACV)*, pages 372–380. IEEE, 2018. 1, 2, 6

[44] Matteo Pagliardini, Prakhar Gupta, and Martin Jaggi. Unsupervised Learning of Sentence Embeddings using Compositional n-Gram Features. In *NAACL 2018 - Conference of the North American Chapter of the Association for Computational Linguistics*, 2018. 8

[45] Jeffrey Pennington, Richard Socher, and Christopher D Manning. Glove: Global vectors for word representation. In *Proceedings of the 2014 conference on empirical methods in natural language processing (EMNLP)*, pages 1532–1543, 2014. 8

[46] AJ Piergiovanni and Michael S Ryoo. Learning shared multimodal embeddings with unpaired data. *CoRR*, 2018. 1

[47] Farhad Pourpanah, Moloud Abdar, Yuxuan Luo, Xinlei Zhou, Ran Wang, Chee Peng Lim, and Xi-Zhao Wang. A review of generalized zero-shot learning methods. *arXiv preprint arXiv:2011.08641*, 2020. 2

[48] Zhaofan Qiu, Ting Yao, and Tao Mei. Learning spatio-temporal representation with pseudo-3d residual networks. In *proceedings of the IEEE International Conference on Computer Vision*, pages 5533–5541, 2017. 2

[49] Alec Radford, Jong Wook Kim, Chris Hallacy, Aditya Ramesh, Gabriel Goh, Sandhini Agarwal, Girish Sastry, Amanda Askell, Pamela Mishkin, Jack Clark, et al. Learning transferable visual models from natural language supervision. *Image*, 2:T2, 2021. 3

[50] Milos Radovanovic, Alexandros Nanopoulos, and Mirjana Ivanovic. Hubs in space: Popular nearest neighbors in high-dimensional data. *Journal of Machine Learning Research*, 11(sept):2487–2531, 2010. 2, 15

[51] Aditya Ramesh, Mikhail Pavlov, Gabriel Goh, Scott Gray, Chelsea Voss, Alec Radford, Mark Chen, and Ilya Sutskever. Zero-shot text-to-image generation. *arXiv preprint arXiv:2102.12092*, 2021. 3

[52] Alina Roitberg, Manuel Martinez, Monica Haurilet, and Rainer Stiefelhagen. Towards a fair evaluation of zero-shot action recognition using external data. In *Proceedings of the European Conference on Computer Vision (ECCV) Workshops*, pages 0–0, 2018. 1, 5

[53] Michael S Ryoo, AJ Piergiovanni, Juhana Kangaspunta, and Anelia Angelova. Assemblenet++: Assembling modality representations via attention connections. In *European Conference on Computer Vision*, pages 654–671. Springer, 2020. 1
10

[54] Yutaro Shigeto, Ikumi Suzuki, Kazuo Hara, Masashi Shimbo, and Yuji Matsumoto. Ridge regression, hubness, and zero-shot learning. In *Joint European conference on machine learning and knowledge discovery in databases*, pages 135–151. Springer, 2015. 15

[55] Khurram Soomro, Amir Roshan Zamir, and Mubarak Shah. Ucf101: A dataset of 101 human actions classes from videos in the wild. *arXiv preprint arXiv:1212.0402*, 2012. 1, 2, 6, 17

[56] Chen Sun, Austin Myers, Carl Vondrick, Kevin Murphy, and Cordelia Schmid. Videobert: A joint model for video and language representation learning. In *Proceedings of the IEEE/CVF International Conference on Computer Vision*, pages 7464–7473, 2019. 3

[57] Tristan Sylvain, Linda Petrini, and Devon Hjelm. Locality and compositionality in zero-shot learning. In *International Conference on Learning Representations*, 2020. 1

[58] Christian Szegedy, Wei Liu, Yangqing Jia, Pierre Sermanet, Scott Reed, Dragomir Anguelov, Dumitru Erhan, Vincent Vanhoucke, and Andrew Rabinovich. Going deeper with convolutions. In *Proceedings of the IEEE conference on computer vision and pattern recognition*, pages 1–9, 2015. 6

[59] Hugo Touvron, Matthieu Cord, Matthijs Douze, Francisco Massa, Alexandre Sablayrolles, and Hervé Jégou. Training data-efficient image transformers & distillation through attention. *arXiv preprint arXiv:2012.12877*, 2020. 3

[60] Du Tran, Lubomir Bourdev, Rob Fergus, Lorenzo Torresani, and Manohar Paluri. Learning spatiotemporal features with 3d convolutional networks. In *Proceedings of the IEEE international conference on computer vision*, pages 4489–4497, 2015. 1, 6

[61] Du Tran, Heng Wang, Lorenzo Torresani, Jamie Ray, Yann LeCun, and Manohar Paluri. A closer look at spatiotemporal convolutions for action recognition. In *Proceedings of the IEEE conference on Computer Vision and Pattern Recognition*, pages 6450–6459, 2018. 1, 2

[62] Ashish Vaswani, Noam Shazeer, Niki Parmar, Jakob Uszkoreit, Llion Jones, Aidan N Gomez, Lukasz Kaiser, and Illia Polosukhin. Attention is all you need. *arXiv preprint arXiv:1706.03762*, 2017. 12

[63] Limin Wang, Yuanjun Xiong, Zhe Wang, Yu Qiao, Dahua Lin, Xiaoou Tang, and Luc Van Gool. Temporal segment networks: Towards good practices for deep action recognition. In *European conference on computer vision*, pages 20–36. Springer, 2016. 4

[64] Qian Wang and Ke Chen. Alternative semantic representations for zero-shot human action recognition. In *Joint European Conference on Machine Learning and Knowledge Discovery in Databases*, pages 87–102. Springer, 2017. 2

[65] Qian Wang and Ke Chen. Zero-shot visual recognition via bidirectional latent embedding. *International Journal of Computer Vision*, 124(3):356–383, 2017. 1, 2

[66] Qian Wang and Ke Chen. Multi-label zero-shot human action recognition via joint latent ranking embedding. *Neural Networks*, 122:1–23, 2020. 2

[67] Zuxuan Wu, Yanwei Fu, Yu-Gang Jiang, and Leonid Sigal. Harnessing object and scene semantics for large-scale video understanding. In *Proceedings of the IEEE Conference on Computer Vision and Pattern Recognition*, pages 3112–3121, 2016. 2

[68] Xiang Xiang, Ye Tian, Austin Reiter, Gregory D Hager, and Trac D Tran. S3d: Stacking segmental p3d for action quality assessment. In *2018 25th IEEE International Conference on Image Processing (ICIP)*, pages 928–932. IEEE, 2018. 1

[69] Xun Xu, Timothy Hospedales, and Shaogang Gong. Semantic embedding space for zero-shot action recognition. In *2015 IEEE International Conference on Image Processing (ICIP)*, pages 63–67. IEEE, 2015. 2

[70] Xun Xu, Timothy Hospedales, and Shaogang Gong. Transductive zero-shot action recognition by word-vector embedding. *International Journal of Computer Vision*, 123(3):309–333, 2017. 2

[71] Xun Xu, Timothy M Hospedales, and Shaogang Gong. Multi-task zero-shot action recognition with prioritised data augmentation. In *European Conference on Computer Vision*, pages 343–359. Springer, 2016. 2

[72] Chenrui Zhang and Yuxin Peng. Visual data synthesis via gan for zero-shot video classification. In *Proceedings of the 27th International Joint Conference on Artificial Intelligence*, pages 1128–1134, 2018. 2, 3

[73] Li Zhang, Tao Xiang, and Shaogang Gong. Learning a deep embedding model for zero-shot learning. In *Proceedings of the IEEE conference on computer vision and pattern recognition*, pages 2021–2030, 2017. 2, 7

[74] Ziming Zhang and Venkatesh Saligrama. Zero-shot learning via joint latent similarity embedding. In *Proceedings of the IEEE Conference on Computer Vision and Pattern Recognition (CVPR)*, June 2016. 5

[75] Xizhou Zhu, Weijie Su, Lewei Lu, Bin Li, Xiaogang Wang, and Jifeng Dai. Deformable detr: Deformable transformers for end-to-end object detection. *arXiv preprint arXiv:2010.04159*, 2020. 3

[76] Yi Zhu, Yang Long, Yu Guan, Shawn Newsam, and Ling Shao. Towards universal representation for unseen action recognition. In *Proceedings of the IEEE conference on computer vision and pattern recognition*, pages 9436–9445, 2018. 2

# Supplementary Material

This appendix presents more details of the limitations, implementation details, and extends the experimental section presented in the main manuscript.

1. **Limitations.** In Section A, we discuss the limitations of the proposed method.
2. **Potential Negative Societal Impact.** We discuss the potential negative societal impact in Section B.
3. **Implementation Details.** We provide network architecture of ResT, dataset statistics, and evaluation protocols in Section C.
4. **Additional Ablation Study.** Additional ablation studies are presented in Section D.
5. **Extended Experiments**. In Section E, we extend the experimental section presented in the main manuscript. We demonstrate the generalization of the proposed method using pre-trained object features as model inputs and visualize qualitative results.

## A. Limitations

The goal of this work is to provide a cleaner framework for zero-shot action recognition. In our setting, the model is not allowed to be pretrained on another dataset, and is evaluated on its ability to perform classification using unseen visual prototypes composited from seen visual prototypes. Although our method demonstrates its competitiveness on the benchmark datasets, it has limitations in some cases. For example, our model confuses similar actions, such as "laugh," "smile," and "chew." In these three classes, the actions mainly involve opening and shutting the jaws, but the muscle movements involved are subtle.

We also observe composite failures, e.g., for "hula hoop" where the class is named only by a noun of the main object, or for "playing daf" where the class is named by a general verb (ex: play, make, use) with a noun of a rare object. Our model is able to composite actions from other actions, but it exists a natural challenge to find relatedness for compositing out-of-distribution objects. However, if the setting of pure zero-shot is relaxed, our model could extend its capability via pretraining on another dataset, such as ImageNet.

Figure 5, 6, and 7 illustrate the confusion matrices of ResT_18 model evaluated on UCF101, HMDB51, and ActivtyNet.

## B. Potential Negative Societal Impact

Training and evaluating video understanding models are typically computationally intensive, which might significantly impact the environment. To alleviate this problem, we proposed a framework that reduces the computational demands for ZSAR. The potential negative impacts may include but are not limited to: (1) It poses a risk when directly applying action recognition models for decision making, especially in the health care and autonomous vehicle fields. (2) A video action recognition model can be misused, for example for unauthorized surveillance. Ethical considerations must be addressed in a real-world application.

## C. Implementation Details

### C.1. Network architecture of ResT

We describe the detailed network architecture of ResT in this section. ResT follows the design of the transformer encoder in [62]. As shown in Figure 8, the transformer encoder consists of alternating layers of multiheaded self-attention (MSA) and MLP blocks (Eq. 7, 8). Residual connections and layernorm (LN) are applied after every block. The MLP contains two fully-connected layers with a GeLU non-linearity. Our transformer consists of L layers. We denote $z_l$ as the output of $l^{\text{th}}$ layers.

$$z'_l = LN(MSA(z_{l-1}) + z_{l-1}), \qquad (7)$$

$$z_l = LN(MLP(z'_l) + z'_l), \qquad (8)$$

where $l = 1, ..., L$.

ResT uses the first token, $z_0^0$, to perform action classification on a source dataset. A classification head is attached to the output of the first token, $z_L^0$. We append a 1-hidden-layer MLP $f(\cdot)$, which is used to predict the final video classes.

$$x = LN(z_L^0) \qquad (9)$$

Our ResT consists of 12 transformer layers with a hidden size of 768D. The visual representation size is also 768D. The classifier weights are $768 \times 664$, and $768 \times 605$ corresponding to Kinetics 664 and 605 training sets.

### C.2. Datasets

We train our models on a subset of the Kinetics dataset [28], and perform evaluations on three action recognition datasets: UCF101, HMDB51, and ActivityNet. UCF101 is labeled with 101 action categories with a focus on sports and contains 13,320 videos. HMDB51 has 6,767 videos with 51 classes. ActivityNet contains 200 classes and 27,801 untrimmed videos with an emphasis on daily activities. Kinetics dataset contains 700 classes with 545,317 training videos.

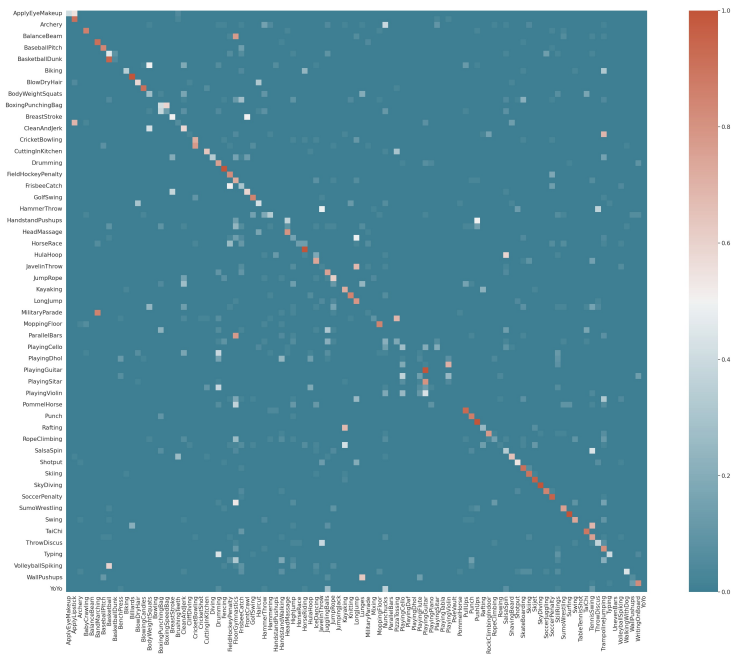

Figure 5. **Confusion matrix on UCF101 by ResT_18 (K664) model.**

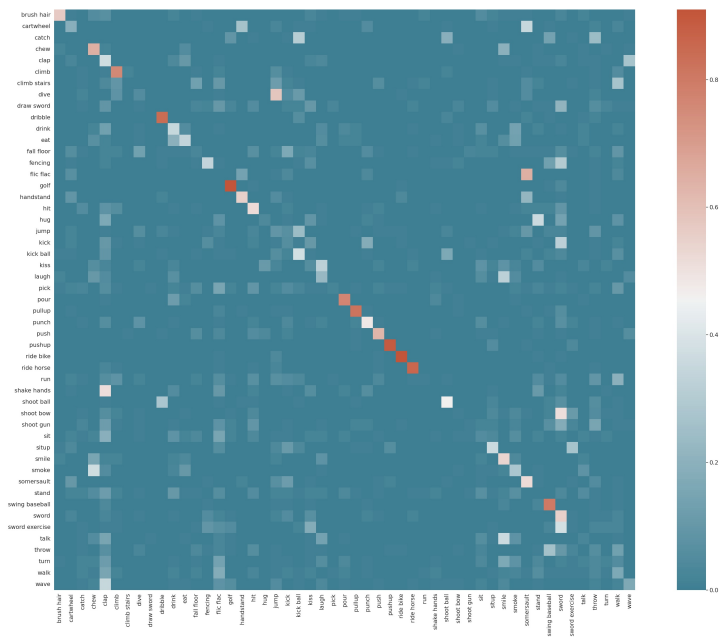

Figure 6. **Confusion matrix on HMDB51 by ResT_18 (K664) model.**

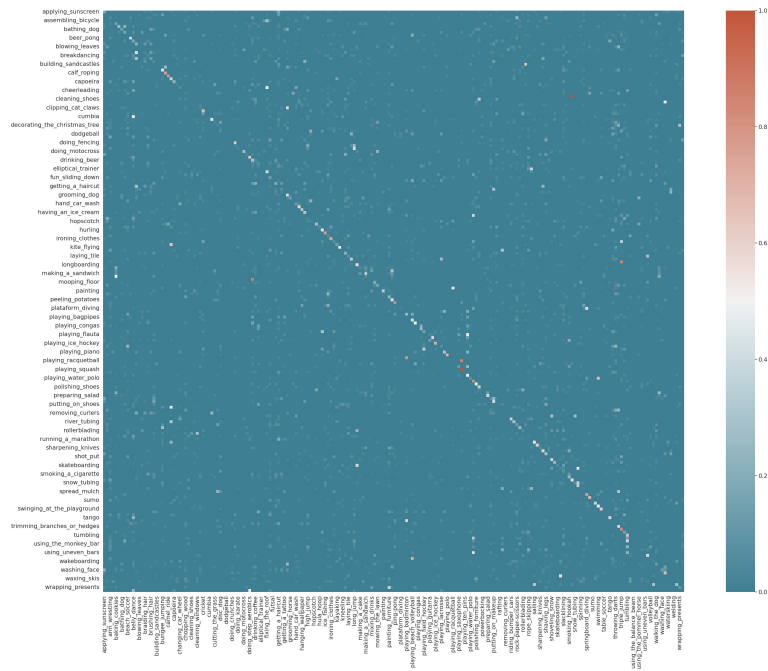

Figure 7. **Confusion matrix on ActivityNet by ResT_18 (K605) model.**

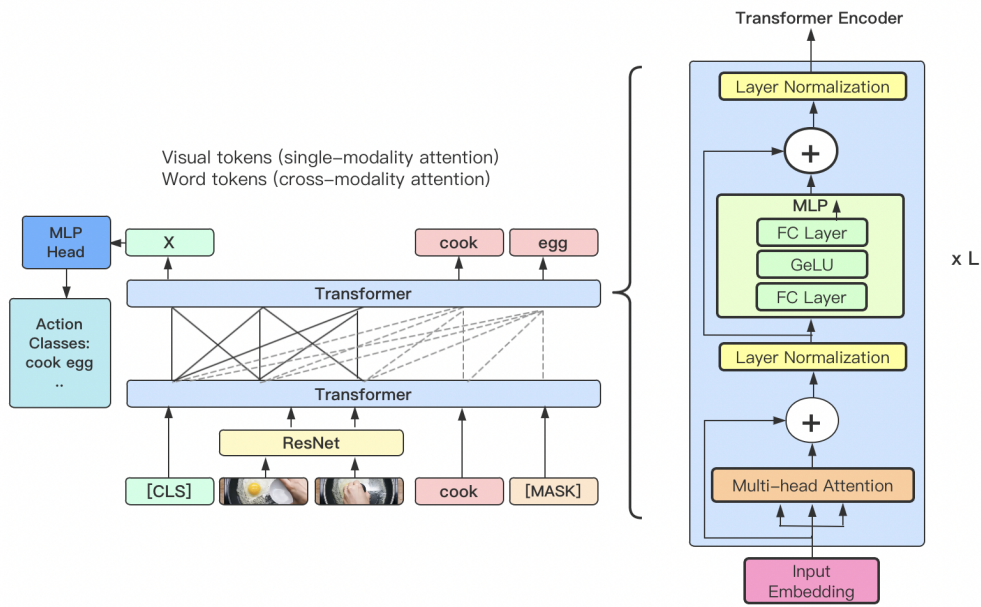

Figure 8. **Architecture of the transformer encoder in our proposed ResT.**
14

## C.3. Evaluation protocol

In the zero-shot evaluation, we report results on half dataset (0/50 split) and full dataset (0/100). Most prior methods use pre-trained action recognition models to extract features, followed by training a ZSL model on 50% of the target dataset and testing on the other 50% of the same dataset to alleviate domain shift problems (50/50 setting). Our work follows E2E [5] to adopt a cross dataset configuration, where the models are only trained once on a source action recognition dataset and then are directly evaluated on 50% of other target datasets. The goal of 0/50 setting is to disallow tailoring ZSAR models to a specific test dataset. In the 0/50 split, we randomly choose 50% classes from the test dataset: 50 on UCF101, 25 on HMDB51, and 100 on ActivityNet. On each test set, we randomly generate 10 splits and report the averaged results. As E2E [5] and our method are trained on a separate dataset, we are able to test our models on full UCF101, HMDB51, and ActivityNet datasets (0/100).

## D. Additional Ablations

In this section, we extend the experimental section presented in the main manuscript.

### D.1. Influence of removing overlapping classes

Table 7. **Accuracy comparisons on models trained on Kinetics 664 and full Kinetics 400/700 datasets.** All models are evaluated on 25 clips on the 50% of UCF101 and HMDB51 datasets.

| ResT_101 Model | UCF (0/50) | |
|---|---|---|
| | Top-1 | Top-5 |
| 664 classes | 58.7 | 75.9 |
| 400 classes | 61.1 | 79.2 |
| 700 classes | 69.2 | 83.8 |

In this experiment, we compare our models trained on Kinetics 664 (with overlapping classes removed) with the models trained on the full Kinetics 400 and 700 datasets [28] (without overlapping classes removed) to demonstrate that removing overlapping classes is a non-trivial learning constraint. The results are reported in Table 7. It can be seen that the models trained on the full Kinetics dataset obtain higher Top-1 accuracy than the models trained on the sets without overlapping classes (e.g., 2.4% absolute gains in Top-1 accuracy on 0/50 configuration from K664 to K400 and 10.5% gains from K664 to K700). As discussed in the main manuscript, one has to ensure that the seen and unseen classes are disjoint and the zero-shot setting is maintained when external datasets are involved.

## D.2. Importance of constraints in semantic relatedness transfer

The design of the transfer scheme aims to embed a combination of the most representative and distinctive information for effective knowledge transfer. It follows, the proposed framework is thus less prone to the hubness problem and the bias with NN search. In this section, we discuss the importance of the constraints in the semantic relatedness transfer.

The hubness problem is related to the high-dimensional nearest neighbor search. That is, some points (hubs) frequently occur in the $k$-nearest neighbor set of other points. The skewness of an empirical $\psi_k$ distribution is typically used to measure the degree of hubness [50, 54]. The distribution $\psi_k$ is the distribution of the number of times ($\psi_k(j)$) the $j^{\text{th}}$ prototype is in the top $k$ nearest neighbors of the test samples. The skewness of the distribution is defined as:

$$\psi_{k\_skewness} = \frac{\sum_{j=1}^{\gamma}(\psi_k(j) - E[\psi_k])^3}{Var[\psi_k]^{\frac{3}{2}}}, \quad (10)$$

where $\gamma$ is the total number of test prototypes. A higher skewness value indicates a more severe hubness issue.

Here, we summarize the three constraints imposed in the semantic relatedness transfer. Constraint I is to ensure the composited unseen prototypes are representative. Constraint II and III together promote the composited visual prototypes of unseen classes to be distinctive from one another.

In Table 8, we discuss the effect of the constraints in terms of the degree of hubness ($\psi_{1\_skewness}$) and classification accuracy (Top-1/ Top-5). The accuracy is evaluated using one clip with ResT_18 (664/605) model. We consider five combinations: (1) Reverse transfer direction (composite semantic representation and perform ZSAR in the semantic space), (2) No constraints imposed, (3) Only constraint I, (4) Only constraint II and III, (5) All constraints.

We draw several conclusions from Table 8: (1) In general, we observe Top-1 accuracy is negatively affected by the presence of hubs, and the hubness problem is more likely to arise in the semantic space than the visual space. We visualize some qualitative results of ZSAR in different spaces in Figure 9. (2) Compared to 'No constraint,' all combinations of the constraints help improve the classification accuracy and alleviate the effect of the hubness problem. (3) When applying constraint I only, Top-5 accuracy is consistently higher because the constraint filters out the less related classes. (4) Constraint II and III are effective, ensuring the distinction of the composited unseen prototypes. In general, it obtains a low hubness value with these two constraints. (5) Combining all three constraints yields a filter-and-refine methodology. Overall, it achieves the best Top-1 accuracy with a relatively low degree of hubness because

Table 8. **Effect of constraints in the semantic relatedness transfer scheme**

(a) Evaluation on UCF101 dataset

|  |  |  | UCF (0/50) | |
|---|---|---|---|---|
| ResT_18 (664) | ZSAR in V or S space | Skewness | Top-1 | Top-5 |
| Reverse transfer | S | 3.350 | 36.3 | 69.2 |
| No constraint | V | 2.235 | 38.2 | 73.9 |
| Constraint I | V | 1.342 | 50.7 | 81.5 |
| Constraint II + III | V | 1.290 | 51.8 | 74.4 |
| All constraints | V | 1.228 | 54.0 | 74.6 |

(b) Evaluation on HMDB51 dataset

|  |  |  | HMDB (0/50) | |
|---|---|---|---|---|
| ResT_18 (664) | ZSAR in V or S space | Skewness | Top-1 | Top-5 |
| Reverse transfer | S | 3.712 | 35.0 | 63.9 |
| No constraint | V | 1.688 | 35.1 | 64.7 |
| Constraint I | V | 1.379 | 37.9 | 68.5 |
| Constraint II + III | V | 0.849 | 38.1 | 64.5 |
| All constraints | V | 1.418 | 39.2 | 66.9 |

(c) Evaluation on ActivityNet dataset

|  |  |  | ActivityNet (0/50) | |
|---|---|---|---|---|
| ResT_18 (605) | ZSAR in V or S space | Skewness | Top-1 | Top-5 |
| Reverse transfer | S | 2.742 | 21.9 | 40.1 |
| No constraint | V | 1.827 | 21.2 | 45.9 |
| Constraint I | V | 1.173 | 25.1 | 51.5 |
| Constraint II + III | V | 1.228 | 25.4 | 40.8 |
| All constraints | V | 1.020 | 26.2 | 47.4 |

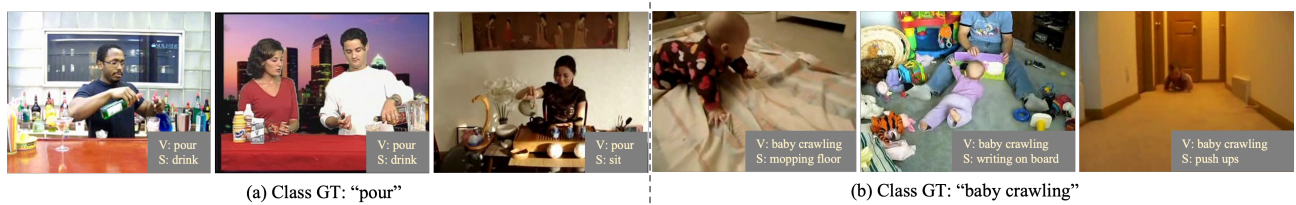

(a) Class GT: "pour"     (b) Class GT: "baby crawling"

Figure 9. **Sample results of ZSAR in visual space (V) and semantic space (S).**

these three constraints together consider both representative and distinctive.

# E. Extended Experiments

Although we propose a framework where no pre-training on additional datasets is performed to ensure no prior knowledge of unseen classes is acquired during training, our model is flexible and capable of cooperating with existing pre-trained models.

In the ablation study, we show the generalization of the proposed model by taking pre-trained object region features as inputs. Considering the essence of zero-shot setting, it might be arguable if using object information is incongruous with the idea of pure ZSAR because it is highly likely that some major objects occur in seen and unseen classes, and some unseen classes are simply named for objects (e.g., "yo-yo," "uneven bars," and "pommel horse"). However, modeling objects helps with the model interpretability. In this experiment, to prevent the model from achieving high accuracy by matching object names instead of recognizing
16

actions, we only use object region features as model inputs to examine the capability of the proposed model. **Detection outputs (object labels) are not used in the experiment.**

In this experiment, we replace frame-level features with object region features. We start with object feature extractors, an off-the-shelf detection network, UpDown [3]. The UpDown detector was trained on Visual Genome dataset [32]. For a frame $I^t$ sampled at time $t$ in a video $v$, an amount $r'^t = [r'^t_1, ..., r'^t_{N^t_r}]$ of $N^t_r$ object features are extracted by the detector, where $r'^t_k \in \mathbb{R}^p$ is a $p$-dimension vector. To encode spatiotemporal information, we construct a 7-d vector $s^t_k$ from the region position (normalized four corner coordinates, width, and height) and the frame index (normalized frame index offset). We concatenate object feature $r'^t_k$ and the spatiotemporal vector $s^t_k$ in order to form a spatiotemporally sensitive region vector $r^t_k$.

We report the results of our model using object region features as model inputs in Table 9. It shows that our model is able to handle contextual information in object features and make the classification of actions relatively effective.

Table 9. **ZSAR performance with ResNet and object features on the 50% of UCF101, HMDB51, and ActivityNet datasets.**

| Model    | UCF101 | HMDB51 | ActivityNet |
|----------|--------|--------|-------------|
| K664     |        |        |             |
| ResT_18  | 54.7   | 39.3   | -           |
| ResT_101 | 58.7   | 41.1   | -           |
| Ours_obj | 57.3   | 39.6   | -           |
| K605     |        |        |             |
| ResT_18  | 50.9   | 37.6   | 29.2        |
| ResT_101 | 55.9   | 40.8   | 32.5        |
| Ours_obj | 55.0   | 40.5   | 34.2        |

Figure 10, 11, and 12 illustrate snapshots of action samples on the UCF101 [55], HMDB51 [33] and ActivityNet [7] that are correctly classified by our model. Each subfigure presents three sample video frames from one action clip. Each frame highlights the five most attended object regions by our network for action recognition. We observe that our model focuses on the active objects where an action is taking place and attends to the most indicative objects, e.g.,"mop handle and head" and "water bucket" in Figure 10(b) or "pizza dough" in Figure 10(c). For example, in Figure 11(c), a sample from the class "eat" on HMDB51, our model attends to the mouth, spoon, and hand. Similarly, in Figure 12(c), a sample from the class "playing beach volleyball" on ActivityNet, our model focuses on the player who sets the volleyball in the first frame, the player who steps toward the ball and bends the knee in the middle frame, and then the same player jumping and preparing for spiking the ball in the last frame. These examples demonstrate the effectiveness of our transformer-based framework that learns to capture the evolution of human actions by observing the most relevant and visually descriptive objects.

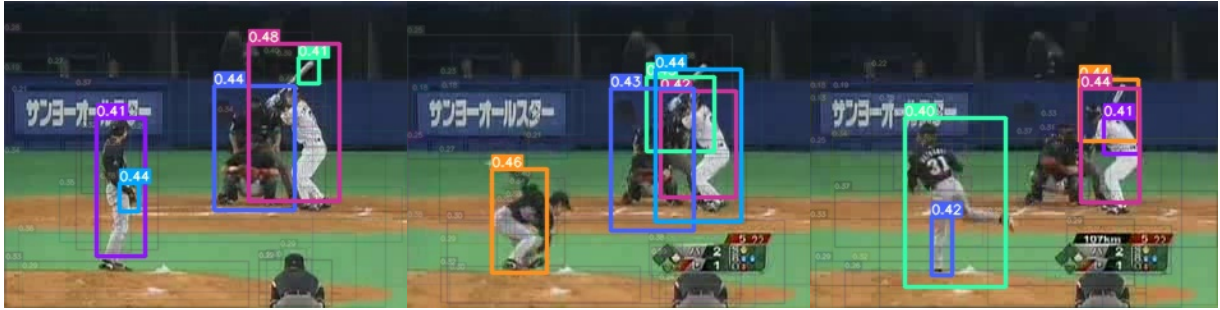

(a) Action class: "**Baseball pitch**".

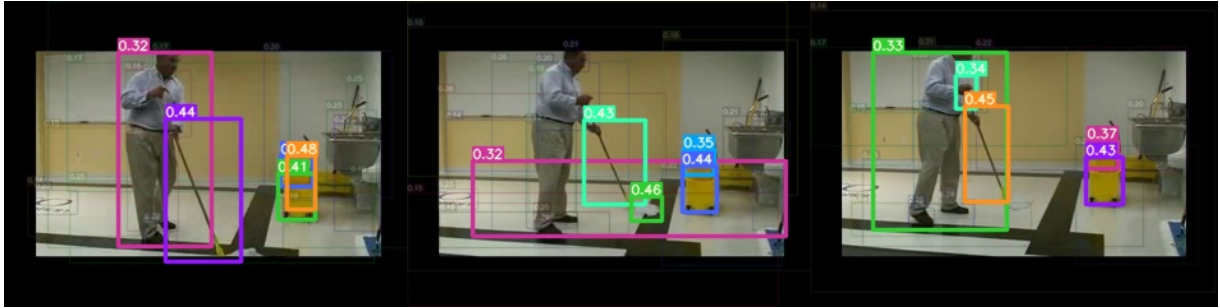

(b) Action class: "**Mopping floor**".

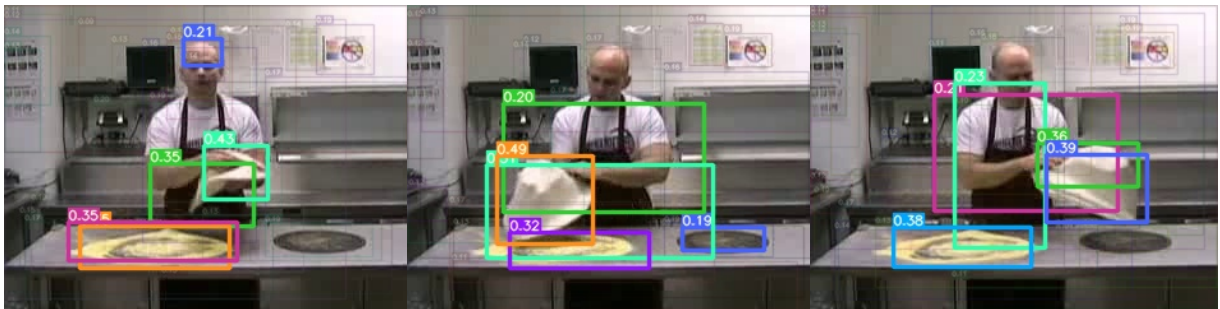

(c) Action class: "**Pizza tossing**".

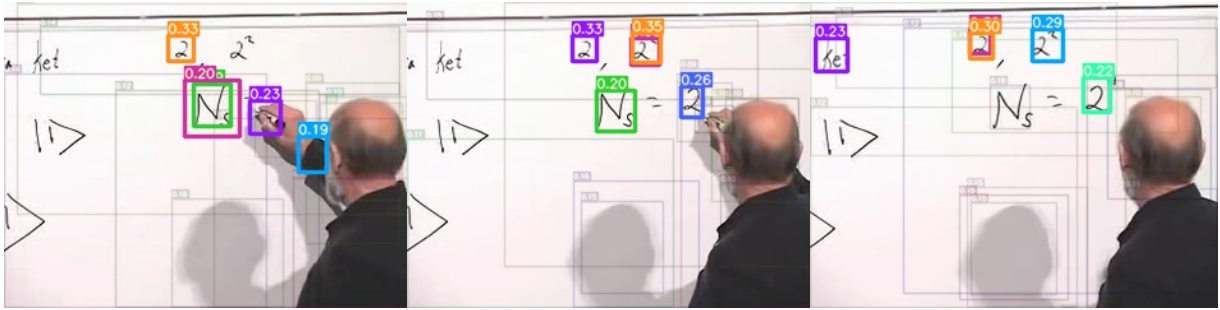

(d) Action class: "**Writing on board**".

Figure 10. Example results from our model with object region features as inputs on the classification of the "baseball pitch," "mopping floor," "pizza tossing," and "writing on board" actions on UCF101 dataset. Different bounding boxes are coded with different colors. Brighter colors depict the most attended objects. Best viewed in color.

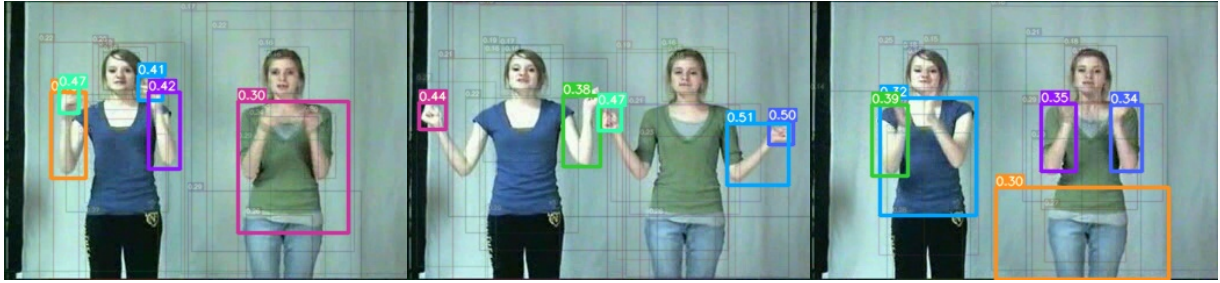
(a) Action class: "**Clap**".

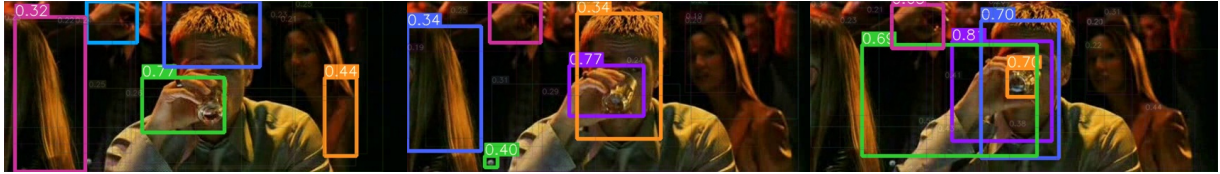
(b) Action class: "**Drink**".

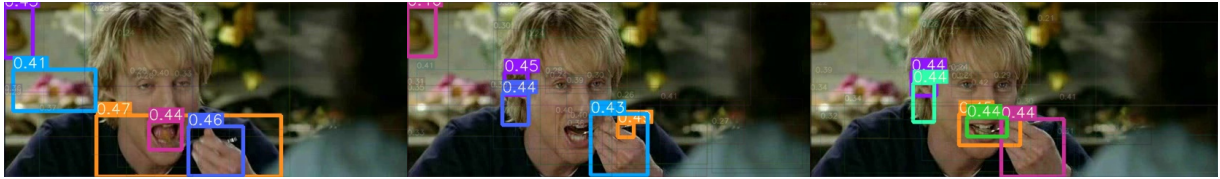
(c) Action class: "**Eat**".

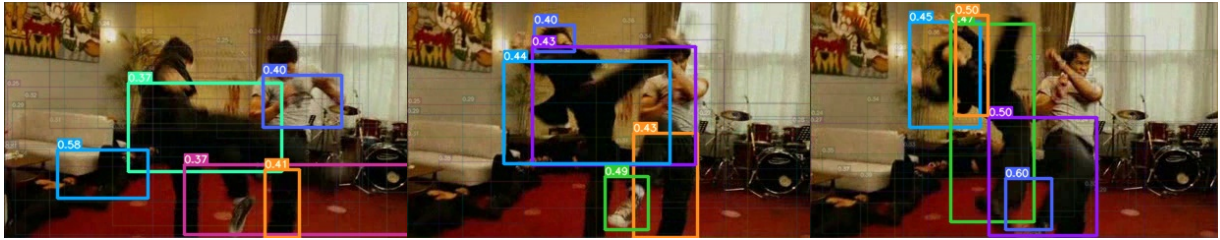
(d) Action class: "**Kick**".

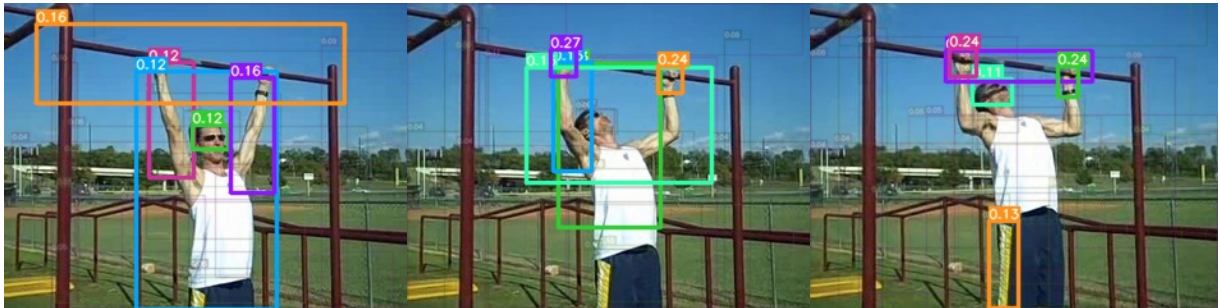
(e) Action class: "**Pullup**".

Figure 11. Example results from our model with object region features as inputs on the classification of the "clap," "drink," "eat," "kick," and "pullup" actions on HMDB51 dataset. Different bounding boxes are coded with different colors. Brighter colors depict the most attended objects. Best viewed in color.

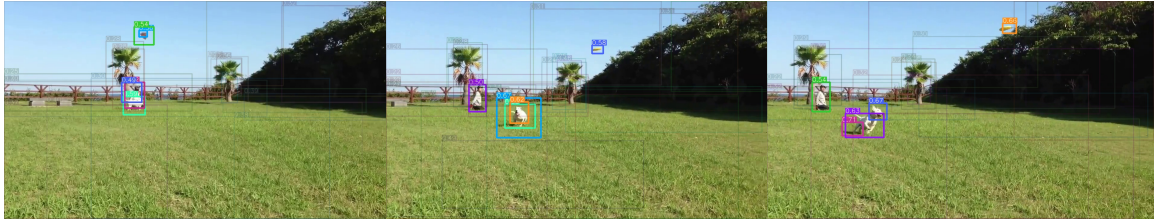
(a) Action class: "**Disc dog**".

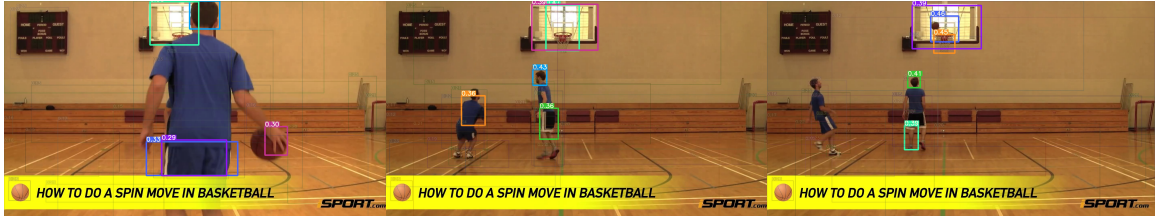
(b) Action class: "**Layup drill in basketball**".

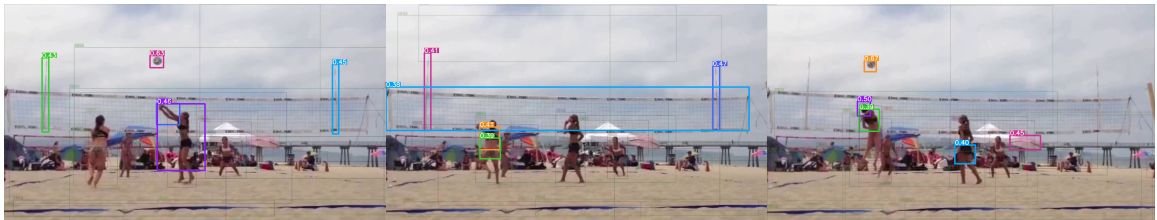
(c) Action class: "**Playing beach volleyball**".

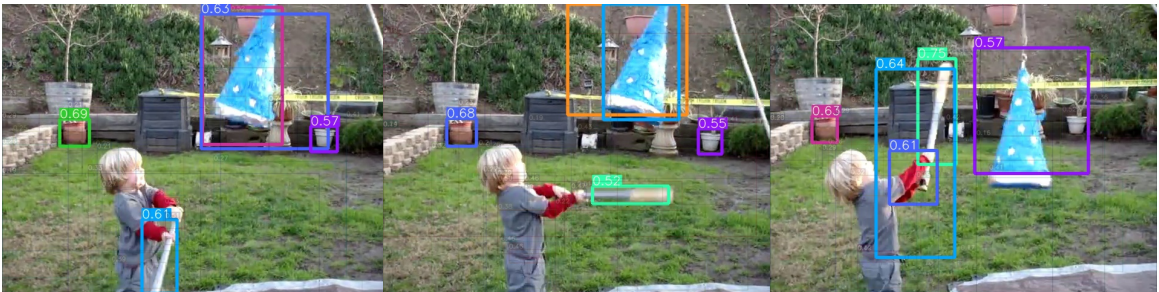
(d) Action class: "**Hitting a pinata**".

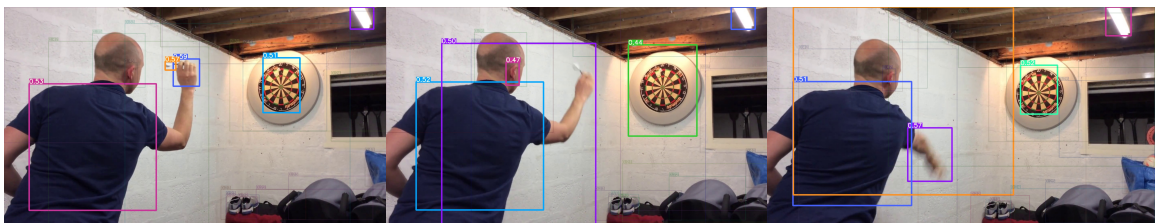
(e) Action class: "**Throwing darts**".

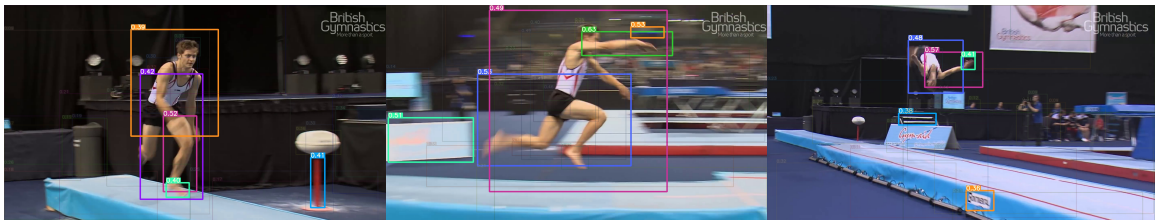
(f) Action class: "**Tumbling**".

Figure 12. Example results from our model with object region features as inputs on the classification of the "disc dog," "layup drill in basketball," "playing beach volleyball," "hitting a pinata," "throwing darts," and "tumbling" actions on ActivityNet dataset. Different bounding boxes are coded with different colors. Brighter colors depict the most attended objects. Best viewed in color.